\documentclass{article}

\PassOptionsToPackage{numbers, compress}{natbib}
\usepackage[main, final]{neurips_2026}

\usepackage[utf8]{inputenc} 
\usepackage[T1]{fontenc}    
\usepackage{hyperref}       
\usepackage{url}            
\usepackage{booktabs}       
\usepackage{amsfonts}       
\usepackage{nicefrac}       
\usepackage{microtype}      
\usepackage{xcolor}         

\usepackage{enumitem} 
\usepackage{xcolor}   
\usepackage[dvipsnames]{xcolor}
\usepackage{booktabs}
\usepackage{amsmath}
\usepackage{makecell}
\usepackage{wrapfig}

\usepackage{multirow}
\usepackage{graphicx} 
\usepackage[table]{xcolor}
\usepackage{tcolorbox}
\usepackage{booktabs}
\usepackage{pifont}
\usepackage{amsthm}
\newtheorem{definition}{Definition}
\newtheorem{proposition}{Proposition}
\newtheorem{lemma}{Lemma}

\title{TrajLift: Lifting Memory Recall Trajectories with Latency-Modulated Heat Diffusion}

\vspace{-3pt}
\author{%
  Jiawen Kang \quad Jinchao Li \quad Mingyu Cui \quad Junan Li \quad Xixin Wu \quad Helen Meng \\
  The Chinese University of Hong Kong, Hong Kong SAR, China \\
  \texttt{jwkang@se.cuhk.edu.hk}
}

\begin{document}

\maketitle

\begin{abstract}
\vspace{-3pt}

Verbal memory retrieval carries rich diagnostic information for Alzheimer's disease through the order, hesitation, and categorical organization of recalled words, yet computational modeling of this retrieval process remains largely unexplored. 
A key challenge is that timing and hierarchical structure are diagnostically entangled: the diagnostic meaning of retrieval timing depends on its hierarchical context.
We propose TrajLift, a spectral framework that models verbal recall as timed trajectories on a hierarchical graph. 
Our central mechanism is latency-modulated heat diffusion on the semantic hierarchy, a unified operator that lifts each recalled word into a representation of both multi-scale hierarchical structure and continuous retrieval timing.
Formal analysis shows that the resulting representation is first-order separable into structural and temporal components with spectral selectivity across hierarchical levels. 
On synthetic benchmarks requiring structure-time interaction, TrajLift stays near-perfect on deep hierarchies where all baselines fail.
On real-world clinical data, we show that TrajLift outperforms diverse baselines by jointly modeling timing and semantic structure, with gains that grow as retrieval timing becomes more informative.
Code is available at \url{https://github.com/kjw11/TrajLift}.
\end{abstract}

\section{Introduction}
\vspace{-3pt}

Alzheimer's disease (AD) is a prevalent neurodegenerative disorder in which memory impairment is identified as one of the most reliable early markers~\citep{martin1985nature}.
Verbal memory tests, which assess memory encoding and retrieval of newly learned word lists, 
\hspace{0.05em} have therefore been widely used as a standard AD

\begin{wrapfigure}{r}{0.4\textwidth}
\vspace{-18pt}
\begin{center}
\includegraphics[width=0.4\textwidth]{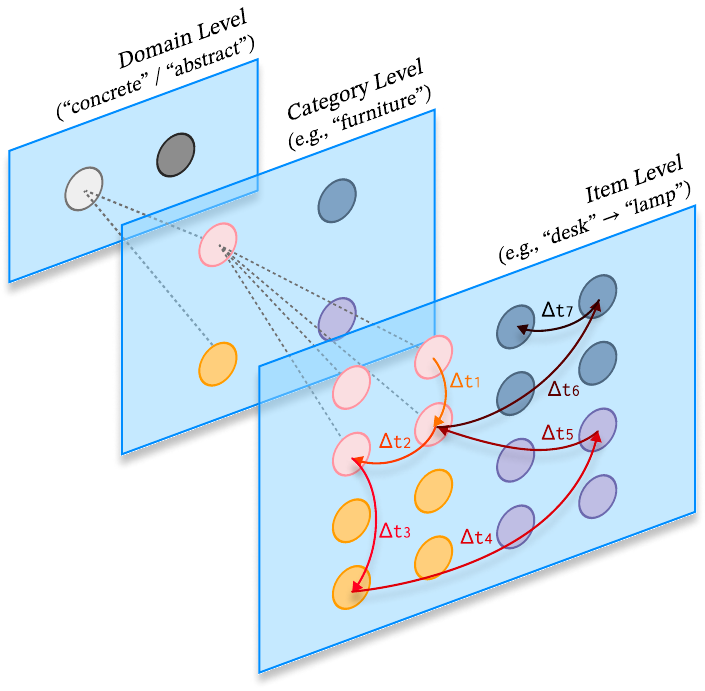}
\end{center}
\vspace{-10pt}
\caption{
\small
Verbal memory retrieval depicted as a timed trajectory on a three-level semantic hierarchy. Arrows indicate recall order; $\Delta t_i$ denotes the latency between consecutive items. 
}
\vspace{-25pt}
\label{fig:highlight}
\end{wrapfigure}
\vspace{-5pt}
screening  protocol~\citep{delis2000cvlt,au2003verbal}.
Clinical and cognitive research has established that the retrieval process 
carries rich diagnostic 
information: the order, hesitation, and categorical organization of recalled words reflect how a subject searches through semantic memory~\citep{collins1969retrieval, behrens2018cognitive}, and patients with AD exhibit distinctively altered patterns in this process~\citep{martin1985nature, au2003verbal}.
However, computational modeling of this retrieval process remains largely unexplored. 
This work aims to model the retrieval dynamics as a structured representation for automatic cognitive screening.

In a typical verbal memory test protocol, subjects are presented with a fixed set of words organized under semantic categories (e.g., "desk" and "lamp" under "furniture") and asked to recall as many as possible after a delay.
A subject's recall then produces an ordered sequence of words, each separated by a measurable latency.
We therefore formalize verbal recall as \textit{a timed trajectory on a hierarchical graph}, where nodes are words, edges are weighted by semantic proximity in the hierarchy, and each step records a visited node together with the inter-item latency.
Figure~\ref{fig:highlight} illustrates this representation.

Learning representations from such trajectories requires capturing the interaction between hierarchical structure and the continuous timing of each transition.
For instance, in semantic fluency tasks, pauses within a semantic cluster and pauses at switches between clusters reflect distinct retrieval processes~\citep{gruenewald1980free}, and AD affects them differently~\citep{troger2019exploitation} — illustrating that the diagnostic interpretation of timing depends on hierarchical context.
This setting falls between established modeling paradigms, and the closest existing approaches come from adjacent fields that were designed for different settings. 
For example, dynamic graph models assume an evolving topology (i.e., changing edges), whereas the semantic hierarchy here is static across all subjects; conversely, graph-based trajectory methods operate on static structures but treat timing as discrete or omit it entirely, whereas retrieval latency is inherently continuous.
We provide a detailed comparison in Table~\ref{tab:comparison}.

In this work, we model verbal retrieval as a graph heat diffusion process on the semantic hierarchy, where each recalled word is treated as a localized signal on the graph that spreads to its structural neighborhood.
Heat diffusion on hierarchically weighted graphs has the well-known property that the diffusion time (i.e., a continuous scalar) controls the structural resolution of the resulting representation.
We exploit this to lift each node visit to progressively coarser hierarchical scales, capturing within-category and cross-domain structure. 
We further extend this framework to encode temporal dynamics by letting the observed inter-item latency modulate the diffusion time within a bounded range, so that latency alters the representation at each scale without changing which hierarchical level that scale resolves.
Formal analysis shows that under this formulation, the structural and temporal signals are first-order separable, and that the temporal encoding is sensitive to the hierarchical level.
We call the resulting framework TrajLift and validate it on both synthetic benchmarks that target specific inductive biases and real-world clinical data for cognitive screening.
Our contributions are summarized as follows:
\vspace{-5pt}
\begin{itemize}
    \item We formalize verbal memory retrieval as a timed trajectory on a hierarchical graph and propose TrajLift, a spectral framework grounded in graph heat diffusion that jointly encodes hierarchical structure and continuous timing in a single operator. This formulation also applies to other domains where ordered events unfold on fixed hierarchies (e.g., user clickstreams on content taxonomies).
    \item We provide formal analysis showing that the temporal modulation decouples structural and temporal signals to first order, and the multi-scale temporal encoding is spectrally selective across hierarchical levels.
    
    \item We validate the framework on synthetic benchmarks that isolate specific inductive biases and on a real-world clinical corpus for cognitive screening, where TrajLift outperforms both conventional clinical measures and computational baselines when retrieval timing is informative.

\begin{table*}[t]
\caption{
\small
Inductive biases of existing approaches for trajectory learning on graphs.
}
\label{tab:comparison}
\vspace{-0.5em}
\begin{center}
\begin{small}
\scalebox{0.75}{
\renewcommand{\arraystretch}{1.2}
\begin{tabular}{lccccc}
\toprule
\textbf{Paradigm} & \textbf{Representative} & \textbf{Static Graph} & \textbf{Hierarchy} & \textbf{Continuous Time} \\
\midrule
Time-Aware Sequence & Time-LSTM~\citep{time-lstm} & - & \ding{55} & \ding{51} \\
Dynamic Graph Learning & TGN~\citep{rossi2020temporal}, TGAT~\citep{xu2020inductive} & \ding{55} & \ding{55} & \ding{51} \\
Session/Trajectory GNN & SR-GNN~\citep{SR-GNN}, TrajGAT~\citep{yao2022trajgat} & \ding{51} & \ding{55} & \ding{55} \\
Hierarchical Pooling & DiffPool~\citep{DiffPool}, SAGPool~\citep{SAGPool} & \ding{51} & \ding{51} (learned) & \ding{55} \\
\rowcolor{gray!15}
\textbf{Ours} & \textbf{Proposed} & \ding{51} & \ding{51} (explicit) & \ding{51} \\
\bottomrule
\end{tabular}}
\end{small}
\end{center}
\vskip -0.1in
\vspace{-5pt}
\end{table*}

\end{itemize}

\vspace{-10pt}
\section{Related Work}
\vspace{-5pt}

\subsection{Automatic AD Detection via Verbal Memory}
\vspace{-5pt}

Verbal memory assessments, such as the California Verbal Learning Test (CVLT)~\cite{delis2000cvlt} and Hong Kong List Learning Test (HKLLT)~\cite{chan1998hkllt}, evaluate cognitive decline by requiring subjects to recall word lists across multiple trials.
Recent work has moved beyond total scores to process-level measures of recall~\cite{gasparini2024normative}, however, such measures, like text-based encoders, reduce a recall sequence to aggregate statistics or a flat string.
They overlook the fundamental nature of verbal retrieval as the navigation of a structured cognitive map~\cite{behrens2018cognitive}, in which the sequence of recalled items forms a specific path through semantic space.
To the best of our knowledge, graph-theoretic modeling of this retrieval path remains underexplored.

\subsection{Inductive Biases in Graph Trajectory Learning}

Modeling retrieval paths requires aligning inductive biases with the cognitive process: a continuous signal evolving over a static hierarchy.
Existing paradigms, however, often misalign with this specific setting.
First, regarding \textit{dynamics}, Continuous-Time Dynamic Graph (CTDG) models~\cite{hosseini2025deep,corradini2025systematic}, such as TGN~\cite{rossi2020temporal} and TGAT~\cite{xu2020inductive}, predominantly focus on \textit{dynamics of graphs} (evolving topology).
This edge-centric bias is suboptimal for our scenario, where the semantic substrate is invariant and the diagnostic signal lies solely in the trajectory of node activations.
Second, regarding \textit{trajectory and hierarchy}, while spatial models like TrajGAT~\cite{yao2022trajgat} and TrajDiff~\cite{zhang2025trajdiff} capture sequential patterns, and encoders like Graphormers~\cite{ying2021transformers,shehzad2026graph} or R-GCNs~\cite{schlichtkrull2018modeling} handle global structure, they typically process transitions as discrete steps.
Consequently, these architectures lack a mechanism to model \textit{continuous} inter-item latencies in relation to the structure in which they occur. They therefore cannot capture how the meaning of a latency (e.g., hesitation) depends on its position in the semantic hierarchy.

\subsection{Graph Spectral Theory}
To address the limitations of discrete message passing, graph spectral theory serves as a fundamental tool for analyzing dynamics on static structures~\cite{fiedler1973algebraic,chung1997spectral,spielman2011spectral}.
By decomposing graph signals into orthogonal frequency components, spectral methods enable the decoupling of local high-frequency fluctuations from global low-frequency structural trends.
Building on this foundation, recent research has evolved from fixed polynomial approximations to \textit{continuous} and \textit{adaptive} spectral filtering~\cite{ding2024survey}.
Notably, continuous diffusion models, such as GRAND~\cite{chamberlain2021grand} and GDC~\cite{gasteiger2019diffusion}, reformulate information propagation as a continuous diffusion process governed by heat or oscillator equations across the graph.
Unlike discrete aggregation, the spectral domain offers a continuous basis for signal evolution, providing a natural way to couple the physical timing of memory retrieval with diffusion on the graph, which underpins our proposed framework.

\section{Method}
\label{sec:method}
 
We formalize the problem of representation learning for timed trajectories on hierarchical graphs and describe TrajLift, a spectral framework that jointly encodes hierarchical structure and retrieval timing through graph heat diffusion.
 
 
\subsection{Problem Setting}
\label{sec:problem}
 
Let $G = (V, W)$ be a weighted undirected graph with node set $V$ of size $N$ and adjacency matrix $W \in \mathbb{R}^{N \times N}$.
The graph remains fixed across all observations and encodes prior knowledge about relationships among nodes.
We consider the case where $V$ is organized into a hierarchical partition with $M$ nested levels.
The following distance measure captures this structure.
 
\begin{definition}[Ultrametric Distance]
\label{def:ultrametric}
Given a hierarchical partition of $V$ with $M$ levels, the ultrametric distance $d_{\mathrm{ultra}}(i, j)$ between nodes $i, j \in V$ is the level of their lowest common ancestor.
\end{definition}
 
\noindent Intuitively, $d_{\mathrm{ultra}} = 1$ indicates sibling nodes sharing the same parent, $d_{\mathrm{ultra}} = 2$ indicates cousins whose parents share a grandparent, and so on.
We use this distance to specify the edge weights of $G$: $W_{ij} = w\big(d_{\mathrm{ultra}}(i,j)\big)$, where $w\colon \{1, \ldots, M\} \to \mathbb{R}^+$ is a monotonically decreasing function.
This assigns stronger connections to hierarchically closer nodes.
In verbal memory assessment, the hierarchy is specified by the test protocol---words are grouped into categories (e.g., ``desk'' and ``lamp'' under ``furniture'') which belong to broader domains---and $G$ remains the same for all subjects.
 
\begin{definition}[Timed Trajectory]
\label{def:trajectory}
A timed trajectory on $G$ is a sequence of timestamped node visits $\sigma = \big((v_1, t_1), (v_2, t_2), \ldots, (v_K, t_K)\big)$, where $v_k \in V$ is the visited node, $t_k \in \mathbb{R}^+$ is the timestamp, and $t_1 < t_2 < \cdots < t_K$.
The \textbf{latency} between consecutive visits is $\delta_k = t_{k+1} - t_k$ for $k = 1, \ldots, K{-}1$.
The trajectory length $K$ may vary across observations and is itself part of the observed behavior.
\end{definition}
 
\noindent\textbf{Goal.}
Given a trajectory $\sigma$ on a hierarchical graph $G$, we aim to learn a representation $\mathbf{z} = f(\sigma, G) \in \mathbb{R}^d$ that captures both structural patterns and temporal dynamics for downstream classification.
In verbal memory assessment, each trajectory records a subject's recall sequence together with inter-item latencies, and the classification target is cognitive status.
 
\subsection{Multi-Scale Structural Representation}
\label{sec:lifting}
 
We now describe how to represent each node visit in a way that reflects its position within the hierarchy.
A visit to node $v_k$ is represented as the indicator vector $\mathbf{x}_k = \mathbf{e}_{v_k} \in \mathbb{R}^N$, which identifies the visited node but carries no information about its hierarchical neighborhood.
To access the hierarchical structure encoded in $G$, below we analyze the graph through its Laplacian.
 
Let $D$ be the degree matrix with $D_{ii} = \sum_j W_{ij}$.
The normalized Laplacian is $L = I - D^{-1/2} W D^{-1/2}$.
Since $L$ is symmetric positive semi-definite, it admits an eigendecomposition $L = U \Lambda U^\top$, where $\Lambda = \mathrm{diag}(\lambda_1, \ldots, \lambda_N)$ with $0 = \lambda_1 \le \cdots \le \lambda_N$, and the columns of $U$ are orthonormal eigenvectors.
These eigenvectors form a spectral basis analogous to Fourier modes: those associated with small eigenvalues vary smoothly across the graph, while those with large eigenvalues oscillate rapidly.
Because the edge weights of $G$ decrease with ultrametric distance (Definition~\ref{def:ultrametric}), the eigenvalues of $L$ concentrate into bands corresponding to different levels of the hierarchy: low-frequency eigenvectors are smooth over fine-grained clusters and capture coarse partitions, while high-frequency eigenvectors retain local within-cluster variation.
 
\noindent\textbf{Spectral lifting.}
To obtain a representation of $\mathbf{x}_k$ that incorporates hierarchical context, we diffuse it across $G$ via the graph \textit{heat kernel} $H_\tau = e^{-\tau L}$.
The heat kernel attenuates each spectral component by $e^{-\tau \lambda_i}$, acting as a low-pass filter whose cutoff is controlled by a continuous scalar parameter, the \textit{diffusion time}~$\tau$.
This continuous parameterization is essential: unlike discrete-step operators such as polynomial graph filters, it allows the resolution of the hierarchy to be controlled by a single scalar.
The \textit{lifted representation} of a node visit is:
\begin{equation}
    \mathbf{z}_k(\tau) \;=\; H_\tau \, \mathbf{x}_k \;=\; e^{-\tau L} \, \mathbf{x}_k.
    \label{eq:lifting}
\end{equation}
Combined with the banded eigenvalue structure described above, this filtering gives rise to a precise relationship between diffusion time and hierarchical resolution.
This property is well established for graphs with hierarchically structured edge weights; we state it for our construction:
 
\begin{proposition}[Hierarchical resolution via heat diffusion]
\label{prop:resolution}
When the edge-weight ratios between adjacent hierarchy levels are sufficiently separated, the Laplacian spectrum decomposes into bands associated with the hierarchy levels.
Since $e^{-\tau \lambda_i}$ decays faster for larger eigenvalues, increasing $\tau$ selectively suppresses higher-frequency bands, continuously interpolating between fine-grained and coarse representations.
\end{proposition}
\noindent The proposition is stated for a balanced two-level hierarchy. For deeper hierarchies, the same argument applies recursively to each pair of adjacent levels. 
A formal statement and proof are provided in Appendix~\ref{app:proofs}.
 
Given Proposition~\ref{prop:resolution}, evaluating the lifting operator at different diffusion times accesses different hierarchical levels.
We can therefore obtain representations at multiple resolutions by choosing several values of $\tau$.
In our setting, the hierarchy has three levels (items, categories, domains), so we instantiate three scales: $\mathbf{z}_k^{(s)} = H_{\tau_s} \mathbf{x}_k$ for $s \in \{\text{micro}, \text{meso}, \text{macro}\}$, aligned with the three levels respectively.
Micro (small~$\tau$) preserves within-category structure, meso (medium~$\tau$) captures cross-category patterns within a domain, and macro (large~$\tau$) retains only cross-domain structure.
The diffusion times are initialized from the spectral structure of $L$ so that each scale's filter response aligns with a distinct eigenvalue band (Appendix~\ref{appendix:implementation}).
 
At this point, each recall event is represented with hierarchical context at three resolutions.
However, two visits to the same node receive identical representations regardless of timing---a one-second latency and a ten-second latency after recalling ``desk'' produce the same lifted signal.
The next section addresses this limitation.
 
 
\subsection{Temporal Encoding}
\label{sec:temporal}
 
The lifted representations from \S\ref{sec:lifting} sample the continuous diffusion at three fixed diffusion times, each resolving a different hierarchical level.
Because the ultrametric weighting produces well-separated eigenvalue bands (\S\ref{sec:lifting}), the hierarchical resolution at each scale is robust to modulation of the diffusion time: as long as the modulated $\tau$ does not reach the range of an adjacent scale, shifting $\tau$ around a base value alters the within-band attenuation profile but does not change which hierarchical level is resolved. The log-compressed latency modulation bounds $\tau$ to at most three times its base value for 90\% of observed latencies, which keeps the three scales in non-overlapping ranges (Appendix~\ref{app:margin}).
This robustness creates a channel for temporal encoding: we can let the observed latency $\delta_k$ modulate the diffusion time at each scale, encoding temporal information without corrupting the structural representation.
 
To characterize the effect of this modulation, we expand $\mathbf{z}_k(\tau + \Delta\tau)$ to first order.
Using $\frac{\partial}{\partial \tau} H_\tau = -L H_\tau$, we obtain
\begin{equation}
    \mathbf{z}_k(\tau + \Delta\tau)
    \;=\;
    \underbrace{H_\tau \mathbf{x}_k}_{\text{zero-order}}
    \;-\;
    \Delta\tau \cdot
    \underbrace{L \, H_\tau \mathbf{x}_k}_{\text{first-order}}
    \;+\;
    O(\Delta\tau^2).
    \label{eq:taylor}
\end{equation}
The zero-order term is the structural lift $\mathbf{z}_k(\tau)$, invariant to the modulation; the first-order term is proportional to $\Delta\tau$ and mediated by the graph Laplacian $L$, confirming that latency enters through a structure-aware correction while the structural base is preserved.
We formalize this decomposition for reference in the spectral analysis that follows.
 
\begin{lemma}[First-order residual separation]
\label{lem:separation}
By Equation~\eqref{eq:taylor}, the modulated lift decomposes into a zero-order structural term invariant to latency and a first-order latency-induced perturbation mediated by the Laplacian $L$.
The residual does not make time independent of structure; it isolates how latency perturbs the structural lift at a given scale.
\end{lemma}
 
\noindent We define the \emph{modulated diffusion time} at scale $s$ for position $k$ as
\begin{equation}
    \tilde{\tau}_k^{(s)} \;=\; \tau_s^{\mathrm{base}} \cdot \big(1 + \gamma_s \cdot \varphi(\delta_k)\big),
    \label{eq:modulation}
\end{equation}
where $\tau_s^{\mathrm{base}} = \mathrm{softplus}(\beta_s)$ is the learnable base diffusion time at scale $s$, $\gamma_s \ge 0$ controls the modulation strength, and $\varphi(\delta) = \log(1 + \delta)$ compresses the dynamic range of observed latencies (discussed in Appendix~\ref{appendix:implementation}).
Applying the heat kernel at the base and modulated diffusion times yields the \emph{baseline lift} $\mathbf{z}_{\mathrm{base},k}^{(s)} = H_{\tau_s^{\mathrm{base}}} \mathbf{x}_k$ and the \emph{modulated lift} $\mathbf{z}_{\mathrm{mod},k}^{(s)} = H_{\tilde{\tau}_k^{(s)}} \mathbf{x}_k$.
Their difference defines the \emph{diffusion residual}:
\begin{equation}
    \Delta \mathbf{z}_k^{(s)} \;=\; \mathbf{z}_{\mathrm{mod},k}^{(s)} - \mathbf{z}_{\mathrm{base},k}^{(s)}.
    \label{eq:residual}
\end{equation}
In practice we compute this difference exactly; Equation~\eqref{eq:taylor} characterizes its leading-order behavior, providing interpretability rather than serving as a computational approximation.
 
In the eigenbasis of $L$ (\S\ref{sec:lifting}), the first-order residual applies a \emph{spectral response} to each frequency component of the input signal.
 
\begin{proposition}[Band-pass temporal selectivity]
\label{prop:bandpass}
Under the first-order approximation (Lemma~\ref{lem:separation}), the temporal residual at scale $s$ applies the spectral response
\begin{equation}
    -\Delta\tau_k^{(s)} \cdot r_s(\lambda),
    \qquad
    r_s(\lambda) \;=\; \lambda \, e^{-\tau_s^{\mathrm{base}} \lambda},
    \label{eq:bandpass}
\end{equation}
where $\Delta\tau_k^{(s)}$ controls the magnitude of the latency effect and $r_s$ determines which spectral frequencies are most sensitive.
The response $r_s$ is band-pass-shaped: it vanishes at $\lambda = 0$ and as $\lambda \to \infty$, and reaches its maximum at $\lambda^\star = 1 / \tau_s^{\mathrm{base}}$.
\end{proposition}
 
\noindent To verify the peak, differentiate $r_s$ with respect to the Laplacian eigenvalue:
$\frac{\partial}{\partial \lambda}\!\big(\lambda \, e^{-\tau_s \lambda}\big)
= e^{-\tau_s \lambda}(1 - \tau_s \lambda)$,
which vanishes at $\lambda^\star = 1/\tau_s^{\mathrm{base}}$.
Since different scales have different $\tau_s^{\mathrm{base}}$, their spectral responses peak at different eigenvalues: the micro-scale residual is most sensitive to high-frequency within-category modes, while the macro-scale residual responds primarily to low-frequency cross-domain modes.
This means that temporal patterns at different hierarchical levels are captured by different residual channels, enabling the model to distinguish temporal signals that arise at different structural resolutions.
For example, within-category hesitation is captured by a different channel than cross-category navigation delays.
When $G$ exhibits separated eigenvalue bands, choosing diffusion scales whose peaks fall into different bands further limits cross-scale overlap; Appendix~\ref{appendix:implementation} provides a sufficient condition.
 
At each trajectory position, we concatenate the modulated lift and the diffusion residual at each scale with layer normalization:
$\mathbf{z}_k^{(s)} = \big[\mathrm{norm}(\mathbf{z}_{\mathrm{mod},k}^{(s)}) \;\|\; \mathrm{norm}(\Delta\mathbf{z}_k^{(s)})\big] \in \mathbb{R}^{2N}$.
Applying this at all three scales yields $Z^{(\mathrm{micro})}, Z^{(\mathrm{meso})}, Z^{(\mathrm{macro})} \in \mathbb{R}^{K \times 2N}$.
 
\subsection{Trajectory Readout}
\label{sec:readout}
 
The multi-scale representations are fused via a gated identity-residual architecture that progressively incorporates coarser context:
\begin{equation}
\begin{aligned}
    \mathbf{F}_0 &= Z^{(\mathrm{micro})}, \\
    \mathbf{F}_1 &= \mathbf{F}_0 + \sigma(\mathbf{F}_0 W_{\mathrm{meso}}) \odot Z^{(\mathrm{meso})}, \\
    \mathbf{F}   &= \mathbf{F}_1 + \sigma(\mathbf{F}_1 W_{\mathrm{macro}}) \odot Z^{(\mathrm{macro})},
\end{aligned}
\label{eq:fusion}
\end{equation}
where $W_{\mathrm{meso}}, W_{\mathrm{macro}} \in \mathbb{R}^{2N \times 2N}$ are learnable projections and $\sigma(\cdot)$ denotes the sigmoid function.
The micro-scale serves as the identity pathway, preserving item-level dynamics; meso- and macro-scale information is incrementally gated in, mirroring the hierarchy where local patterns provide the foundation and coarser context serves as refinement.
 
The fused representation is projected to the target dimension $\mathbf{E} = \mathbf{F} W_{\mathrm{proj}} + \mathbf{b}_{\mathrm{proj}} \in \mathbb{R}^{K \times D}$.
To obtain a trajectory-level representation, we aggregate the positional embeddings using a two-layer LSTM, which is well suited to the short, variable-length trajectories in our setting.
The resulting representation $\mathbf{h} \in \mathbb{R}^D$ is passed to a linear classifier $\hat{\mathbf{y}} = \mathrm{softmax}(W_c \mathbf{h} + \mathbf{b}_c)$, and the model is trained end-to-end by minimizing cross-entropy loss.
\vspace{-5pt}
\section{Data Description}
\vspace{-5pt}

\subsection{Synthetic Data}
To isolate and validate the necessity of joint structure-time modeling, we design a synthetic benchmark where neither modality alone provides a discriminative signal. We term this the \textit{Structure-Time XOR (ST-XOR)} benchmark.

\begin{table}[h!]
\centering
\vspace{-3pt}
\caption{\small Structural specifications of the hierarchy variants in the ST-XOR benchmark.}
\label{tab:hierarchy_variants}

\scalebox{0.85}{
\begin{tabular}{@{}lcll@{}}
\toprule
\textbf{Variant} & \textbf{Levels} & \textbf{Structure} & \textbf{Cluster Configuration} \\
\midrule

\textit{ST-XOR} 
& 2 
& $16 \to 64$ 
& 16 clusters \\

\textit{ST-XOR-w} 
& 2 
& $32 \to 256$ 
& 32 clusters (\textbf{w}ider) \\

\textit{ST-XOR-wh} 
& 3 
& $4 \to 32 \to 256$ 
& 4 super- / 32 sub- (\textbf{w}ider + \textbf{h}igher) \\

\bottomrule
\end{tabular}
\vspace{-3pt}

}
\end{table}

\noindent\textbf{Graph Construction.} 
We construct hierarchies following the ultrametric substrate with three variants of increasing complexity, as detailed in Table~\ref{tab:hierarchy_variants}.

\noindent\textbf{Trajectory Generation.}
Each trajectory is generated via a sequential sampling process of length 12 from a uniformly sampled leaf node. Each step transitions to an intra-cluster node (same parent) or an inter-cluster node (different parent) with equal probability. Self-loops are permitted to decouple trajectory length from transition count.

\noindent\textbf{Class Design.}
The two classes differ only in timing-structure correlation at a designated hierarchical level:
(a) \textit{Coherent}: fast intra-cluster, slow inter-cluster ($\mu_{\text{fast}}$=0.5s, $\mu_{\text{slow}}$=3.0s);
(b) \textit{Chaotic}: slow intra-cluster, fast inter-cluster.
For ST-XOR and ST-XOR-w, cluster membership is defined at the 16- and 32-node levels, respectively. For ST-XOR-wh, it is defined at the 4-node super-cluster level, requiring models to perceive coarser structural distinctions—a more challenging setting.

\noindent\textbf{XOR Trap.}
This design creates an XOR trap: both classes share identical marginal distributions over edge types (50\% each) and identical mean latencies ($\approx$1.75s), rendering structure-only and time-only models ineffective. Only models capturing the structure-time interaction can distinguish the classes.

\noindent\textbf{Dataset Statistics.}
Each variant comprises 2,000 balanced trajectories (1,000 per class), split into training (70\%), validation (15\%), and test (15\%).

\subsection{Real-world Data}
\vspace{-5pt}
We evaluate on a clinical speech corpus collected for cognitive screening among older adults. Each session includes the Hong Kong List Learning Test (HKLLT), a standardized verbal memory protocol requiring recall of a 16-word list organized under a three-level semantic hierarchy (see Appendix~\ref{app:hkllt}).

\noindent\textbf{Task Selection.} 
HKLLT comprises multiple trials; we focus on Trial 4 (10-minute delayed recall) and Trial 5 (30-minute delayed recall), both clinically sensitive to early-stage AD. We treat these as separate benchmarks, denoted HKLLT-4 and HKLLT-5.

\noindent \textbf{Preprocessing.}
Recordings were manually transcribed with word-level timestamps. We filtered task-irrelevant content (e.g., fillers and off-topic utterances) and retained only valid recalled items, yielding clean timed trajectories.

\noindent \textbf{Participants.}
Inclusion criteria required participants to be aged 60+ with adequate hearing and vision.
Each participant was labeled as normal cognition (NC), mild neurocognitive disorder (NCD), or major NCD by clinical experts. We merged mild and major NCD into a single NCD class.
The corpus contains 402 participants. We exclude 15 participants with annotation issues in Trial 4 or Trial 5, so that the two benchmarks share the same 387 participants (NC = 233, NCD = 154).

\section{Experiments}
\subsection{Experimental Settings}
\textbf{Evaluation Protocol.} 
On the synthetic benchmarks, we report accuracy over five random seeds. 
On HKLLT, we report accuracy, macro-F1, and sensitivity (recall on the NCD class) over 15 stratified 80/20 train/test splits shared across all models.
Scores are reported as mean and standard deviation.
Comparisons with baselines use paired differences over the same splits, with Nadeau--Bengio corrected 95\% confidence intervals~\citep{nadeau2003inference}.
For each type of baseline, we compare against its strongest model.

\textbf{Hyperparameter Selection.} 
In preliminary experiments, we randomly hold out 1/5 of the training set as a validation set to determine hidden dimensions for all tested models. 
For the proposed method, the diffusion times $\tau$ are initialized from the graph spectrum and the modulation coefficients $\gamma$ from training-set latency statistics (Appendix~\ref{appendix:implementation}).
No additional hyperparameter tuning is performed.

\textbf{Trajectory Learning Baselines.} 
To evaluate the representational capacity of our framework, we compare it against four categories of baselines:
(a) time-informed sequential models: LSTM and Transformer with Time2Vec~\citep{kazemi2019time2vec}; 
(b) dynamic graph models: TGN and TGAT; 
(c) time-informed graph models: TrajGAT and Graphormer with Time2Vec; 
(d) time-informed graph + sequential models: GAT~\citep{velivckovic2017graph} or R-GCN~\citep{schlichtkrull2018modeling} combined with LSTM and Time2Vec. 
Category (d) serves as a controlled baseline to isolate the contribution of spectral lifting and temporal modulation.
Baselines augmented with Time2Vec are denoted with a `$+$' superscript.

For hierarchy-aware comparisons, the static-graph baselines in categories (c) and (d) are provided with the same fixed semantic hierarchy used by TrajLift, whereas the dynamic graph baselines in category (b) are kept in their native temporal-edge formulation, as their architectures model evolving event edges rather than a fixed graph (Appendix~\ref{appendix:implementation}).
As no prior work directly addresses this task, these baselines are selected for methodological coverage rather than recency.

\textbf{NCD Detection Baselines.} 
To compare learned representations against conventional approaches, we consider two paradigms:
(a) We employ pretrained language models (Chinese BERT and RoBERTa) to encode recalled word sequences as flat text, followed by a lightweight classifier. These baselines entirely ignore retrieval dynamics. 
(b) Following clinical practice, we train logistic regression on handcrafted measures: recall count alone, or a low-level feature set of recall count, sequence length, category transitions, semantic clustering, mean inter-item latency, age, and years of education. 
The low-level feature set includes demographic inputs that TrajLift does not receive. 
Feature definitions are given in Appendix~C, and a variant with all pairwise interactions in Appendix~D.2.3.

\subsection{Results}

\noindent \textbf{Synthetic Experiments.}
Table~\ref{tab:syn} reports classification accuracy on the ST-XOR benchmark. 
We highlight four findings:
(a) Structure-agnostic models fail uniformly: LSTM$^{+}$ and Transformer$^{+}$ reach at most 68\% on ST-XOR and stay near chance on the wider variants (49--53\%).
Without access to the graph structure, these models cannot distinguish intra-cluster from inter-cluster transitions, rendering the timing signal uninterpretable.
(b) Models without explicit aggregation fail to scale: TGN and Graphormer$^{+}$ attain moderate accuracy on ST-XOR (73--76\%), likely by implicitly memorizing cluster memberships for individual nodes.
However, this strategy breaks down as the node count increases: both degrade to chance level on ST-XOR-w and ST-XOR-wh (49--55\%).
(c) Local aggregation fails on deeper hierarchies: TGAT, TrajGAT$^{+}$, and the hybrid baselines (GAT$^{+}$/R-GCN$^{+}$ with LSTM) achieve strong performance on two-level hierarchies ($>$82\% on ST-XOR and ST-XOR-w), as neighborhood aggregation enables perception of immediate cluster structure.

\begin{table}[t]
\centering
\caption{Classification accuracy (\%) on the ST-XOR synthetic benchmark, mean $\pm$ SD over five random seeds. Accuracies above 90\% are underlined.}
\label{tab:syn}
\setlength{\tabcolsep}{3pt}
\renewcommand{\arraystretch}{1.2}
\newcommand{\datastrut}{\rule[-1.0ex]{0pt}{3.2ex}}
\resizebox{\textwidth}{!}{
\begin{tabular}{@{}l cc cc cc cc c@{}}
\toprule
\multirow{2}{*}[-12pt]{\textbf{Benchmark}} &
\multicolumn{2}{c}{\textit{\begin{tabular}[c]{@{}c@{}}Time-informed\\Sequential Models\end{tabular}}} &
\multicolumn{2}{c}{\textit{\begin{tabular}[c]{@{}c@{}}Dynamic\\Graph Models\end{tabular}}} &
\multicolumn{2}{c}{\textit{\begin{tabular}[c]{@{}c@{}}Time-informed\\Graph Models\end{tabular}}} &
\multicolumn{2}{c}{\textit{\begin{tabular}[c]{@{}c@{}}Time-informed Graph \\+Sequential Models\end{tabular}}} &
\multirow{2}{*}[-12pt]{\begin{tabular}[c]{@{}c@{}}\textbf{TrajLift}\\\textbf{(Ours)}\end{tabular}} \\
\cmidrule(lr){2-3} \cmidrule(lr){4-5} \cmidrule(lr){6-7} \cmidrule(lr){8-9}
 & LSTM$^{+}$ & Transformer$^{+}$ & TGN & TGAT & TrajGAT$^{+}$ & Graphormer$^{+}$ & \begin{tabular}[c]{@{}c@{}}GAT$^{+}$\\w/ LSTM\end{tabular} & \begin{tabular}[c]{@{}c@{}}R-GCN$^{+}$\\w/ LSTM\end{tabular} & \\
\midrule
\datastrut ST-XOR & 67.78\textsubscript{$\pm$0.43} & 54.22\textsubscript{$\pm$4.13} & 75.56\textsubscript{$\pm$5.35} & \underline{99.78}\textsubscript{$\pm$0.16} & \underline{95.89}\textsubscript{$\pm$1.03} & 73.56\textsubscript{$\pm$2.11} & \underline{98.78}\textsubscript{$\pm$0.42} & \underline{99.44}\textsubscript{$\pm$0.31} & \cellcolor{gray!10}\underline{98.00}\textsubscript{$\pm$0.98} \\
\datastrut ST-XOR-w & 52.56\textsubscript{$\pm$0.96} & 50.11\textsubscript{$\pm$2.74} & 51.67\textsubscript{$\pm$2.13} & \underline{91.56}\textsubscript{$\pm$1.10} & 82.78\textsubscript{$\pm$1.34} & 55.00\textsubscript{$\pm$3.34} & \underline{99.33}\textsubscript{$\pm$0.51} & \underline{98.78}\textsubscript{$\pm$0.96} & \cellcolor{gray!10}\underline{99.22}\textsubscript{$\pm$0.85} \\
\datastrut ST-XOR-wh & 49.78\textsubscript{$\pm$0.31} & 49.44\textsubscript{$\pm$2.38} & 49.33\textsubscript{$\pm$2.99} & 47.44\textsubscript{$\pm$2.08} & 49.56\textsubscript{$\pm$0.68} & 50.67\textsubscript{$\pm$1.36} & 48.89\textsubscript{$\pm$1.57} & 49.56\textsubscript{$\pm$0.68} & \cellcolor{gray!10}\underline{98.67}\textsubscript{$\pm$0.57} \\
\bottomrule
\end{tabular}}
\vspace{-5pt}
\end{table}
However, all collapse to chance level on ST-XOR-wh (47--50\%), where discrimination requires perceiving super-cluster relationships that lie beyond their local receptive fields.
(d) TrajLift maintains consistent performance: Our method achieves 98--99\% accuracy across all variants, including ST-XOR-wh, where all baselines fail.
We attribute it to the fact that spectral lifting provides explicit access to multiple hierarchical scales through diffusion time, enabling robust structure-time reasoning on both shallow and deep hierarchies.

\begin{table}[t]
\centering
\caption{NC-vs-NCD classification on HKLLT delayed recall (\%), mean $\pm$ SD over 15 splits. Paired comparisons are shown in Figure~\ref{fig:forest}. Sens.\ = sensitivity. Best result per column in bold.}
\label{tab:main}
\scalebox{0.9}{
\setlength{\tabcolsep}{4pt}
\begin{tabular}{lcccccc}
\toprule
 & \multicolumn{3}{c}{HKLLT-4} & \multicolumn{3}{c}{HKLLT-5} \\
\cmidrule(lr){2-4} \cmidrule(lr){5-7}
Model & Acc & F1 & Sens. & Acc & F1 & Sens. \\
\midrule
\multicolumn{7}{l}{\textit{NCD Detection Baselines}} \\
Recall count (LR) & 74.96\textsubscript{$\pm$4.59} & 74.02\textsubscript{$\pm$4.36} & \textbf{73.55}\textsubscript{$\pm$11.71} & 75.90\textsubscript{$\pm$4.39} & 75.19\textsubscript{$\pm$4.53} & 74.84\textsubscript{$\pm$7.34} \\
Low-level features (LR, 7-D) & 74.79\textsubscript{$\pm$4.07} & 73.99\textsubscript{$\pm$4.28} & 73.12\textsubscript{$\pm$8.93} & 75.90\textsubscript{$\pm$3.97} & 75.07\textsubscript{$\pm$4.00} & 73.33\textsubscript{$\pm$7.26} \\
BERT & 72.05\textsubscript{$\pm$5.10} & 70.00\textsubscript{$\pm$5.20} & 58.06\textsubscript{$\pm$6.90} & 74.19\textsubscript{$\pm$3.94} & 72.33\textsubscript{$\pm$4.32} & 61.29\textsubscript{$\pm$7.42} \\
RoBERTa & 75.56\textsubscript{$\pm$4.56} & 73.51\textsubscript{$\pm$5.17} & 60.97\textsubscript{$\pm$9.35} & 73.76\textsubscript{$\pm$3.77} & 71.74\textsubscript{$\pm$4.30} & 60.11\textsubscript{$\pm$9.03} \\
\midrule
\multicolumn{7}{l}{\textit{Trajectory Learning Baselines}} \\
LSTM$^{+}$ & 73.85\textsubscript{$\pm$4.46} & 71.54\textsubscript{$\pm$5.27} & 58.92\textsubscript{$\pm$11.53} & 73.46\textsubscript{$\pm$4.58} & 71.75\textsubscript{$\pm$4.95} & 62.37\textsubscript{$\pm$8.81} \\
Transformer$^{+}$ & 75.00\textsubscript{$\pm$3.36} & 72.45\textsubscript{$\pm$4.26} & 57.10\textsubscript{$\pm$8.68} & 73.50\textsubscript{$\pm$5.12} & 71.32\textsubscript{$\pm$5.74} & 58.71\textsubscript{$\pm$10.05} \\
TGN & 69.02\textsubscript{$\pm$4.34} & 64.41\textsubscript{$\pm$5.51} & 42.80\textsubscript{$\pm$10.38} & 70.47\textsubscript{$\pm$4.76} & 66.58\textsubscript{$\pm$6.40} & 47.96\textsubscript{$\pm$12.94} \\
TGAT & 67.53\textsubscript{$\pm$5.81} & 63.04\textsubscript{$\pm$6.13} & 47.20\textsubscript{$\pm$7.80} & 68.95\textsubscript{$\pm$4.75} & 63.99\textsubscript{$\pm$5.38} & 47.25\textsubscript{$\pm$8.80} \\
TrajGAT$^{+}$ & 74.62\textsubscript{$\pm$3.01} & 72.01\textsubscript{$\pm$3.64} & 56.13\textsubscript{$\pm$7.35} & 75.13\textsubscript{$\pm$5.24} & 73.63\textsubscript{$\pm$5.48} & 64.95\textsubscript{$\pm$7.89} \\
Graphormer$^{+}$ & 74.57\textsubscript{$\pm$3.94} & 72.43\textsubscript{$\pm$4.57} & 59.78\textsubscript{$\pm$9.59} & 75.56\textsubscript{$\pm$3.42} & 73.79\textsubscript{$\pm$3.94} & 63.66\textsubscript{$\pm$9.28} \\
GAT$^{+}$ w/ LSTM & 74.49\textsubscript{$\pm$4.54} & 72.04\textsubscript{$\pm$5.17} & 57.53\textsubscript{$\pm$9.22} & 73.33\textsubscript{$\pm$4.15} & 71.04\textsubscript{$\pm$5.13} & 58.60\textsubscript{$\pm$11.22} \\
R-GCN$^{+}$ w/ LSTM & 75.77\textsubscript{$\pm$3.76} & 73.67\textsubscript{$\pm$4.30} & 60.54\textsubscript{$\pm$7.95} & 72.99\textsubscript{$\pm$5.14} & 71.09\textsubscript{$\pm$5.72} & 60.32\textsubscript{$\pm$9.10} \\
\midrule
\rowcolor{gray!10}
\textbf{TrajLift (Ours)} & \textbf{77.52}\textsubscript{$\pm$5.63} & \textbf{76.22}\textsubscript{$\pm$5.97} & 69.35\textsubscript{$\pm$10.42} & \textbf{80.38}\textsubscript{$\pm$4.63} & \textbf{79.67}\textsubscript{$\pm$4.66} & \textbf{78.92}\textsubscript{$\pm$8.63} \\
\bottomrule
\end{tabular}
}
\vspace{-15pt}
\end{table}

\noindent\textbf{Real-world Detection.} \hspace{0.03em}
We first consider the \hspace{0.02em} results on \hspace{0.02em} HKLLT-5. \hspace{0.02em}
On this trial, \hspace{0.02em} TrajLift achieves 

\begin{wrapfigure}{r}{0.55\textwidth}
\vspace{-19pt}
\begin{center}
\includegraphics[width=0.55\textwidth]{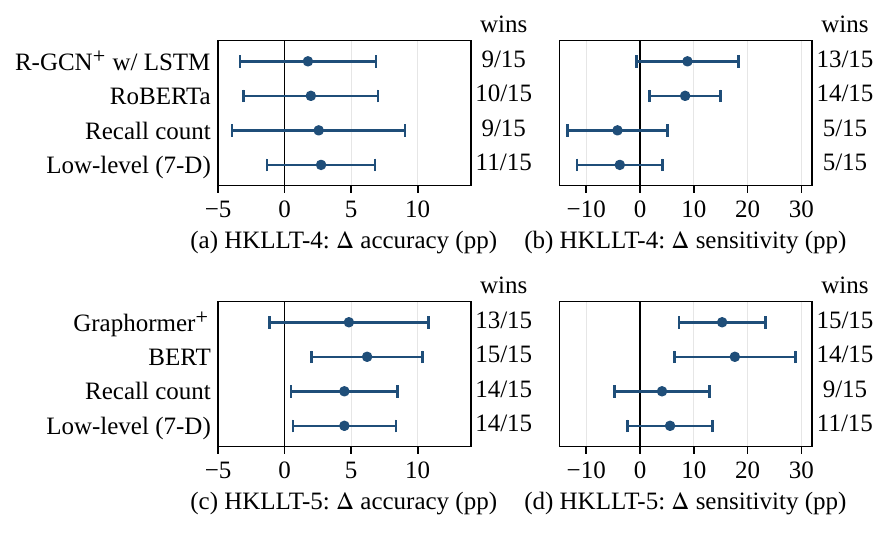}
\end{center}
\vspace{-15pt}
\caption{
\small
Paired differences between TrajLift and selected strong baselines, with 95\% confidence intervals. Wins: splits on which TrajLift is better.
}
\vspace{-20pt}
\label{fig:forest}
\end{wrapfigure}

\vspace{-5pt}
the best accuracy, F1, and sensitivity (Table~\ref{tab:main}), and Figure~\ref{fig:forest} shows that its paired accuracy 
gain is significant over three of the four compared baselines, holding on 13--15 of the 15 splits.
In accuracy, TrajLift outperforms both handcrafted baselines by 4.5 percentage points, whereas adding mean latency, semantic clustering, category transitions, and demographic features to recall count brings no improvement.
Such statistics aggregate over the whole trajectory, while TrajLift encodes each pause in the hierarchical context in which it occurs.
On HKLLT-4, the paired accuracy differences are not significant, despite TrajLift still achieving the highest mean accuracy and F1 and being better on 9--11 of the 15 splits.
Next, we examine why the two trials differ.

\noindent\textbf{Differences Between Trials.}
To understand why TrajLift behaves differently on the two trials, which share the word list, participants, and hierarchy, we compare how strongly each signal separates the NC and NCD groups (Cohen's $|d|$, computed from the statistics in Appendix~C).
Recall count separates the groups similarly in both trials ($1.44$ vs.\ $1.50$) and semantic clustering moderately more on HKLLT-5 ($0.74$ vs.\ $0.93$), whereas the separation by mean inter-item latency roughly doubles ($0.32$ vs.\ $0.64$).
If TrajLift's advantage on HKLLT-5 stems from this timing signal, removing timing information at test time should hurt accuracy more on HKLLT-5 than on HKLLT-4.
Table~\ref{tab:perturb} shows that setting all latencies to zero or to their mean degrades accuracy on both trials, but the drop on HKLLT-5 is about twice as large (6.8 and 17.3 vs.\ 3.9 and 7.2 percentage points), mirroring the difference in latency effect sizes.
These results suggest that TrajLift's advantage scales with the timing signal available in the data: it matches recall count alone when timing carries little additional information and exceeds it when timing becomes more informative.

\begin{table}[t]
\centering
\caption{Test-time latency perturbation (accuracy, \%), mean $\pm$ SD over an independent set of 15 splits. Parentheses: change from the unperturbed input.}
\label{tab:perturb}
\scalebox{0.8}{
\begin{tabular}{lcc}
\toprule
Input latencies & HKLLT-4 & HKLLT-5 \\
\midrule
Original & 77.01\textsubscript{$\pm$5.68} & 80.85\textsubscript{$\pm$4.73} \\
$\delta_k \to 0$ & 73.08\textsubscript{$\pm$4.05} ($-$3.93) & 74.02\textsubscript{$\pm$2.35} ($-$6.84) \\
$\delta_k \to \bar{\delta}$ & 69.83\textsubscript{$\pm$6.94} ($-$7.18) & 63.59\textsubscript{$\pm$5.03} ($-$17.26) \\
\bottomrule
\end{tabular}
}
\vspace{-10pt}
\end{table}
\begin{figure}[h]
\begin{center}
\includegraphics[width=1\linewidth,scale=1.0]{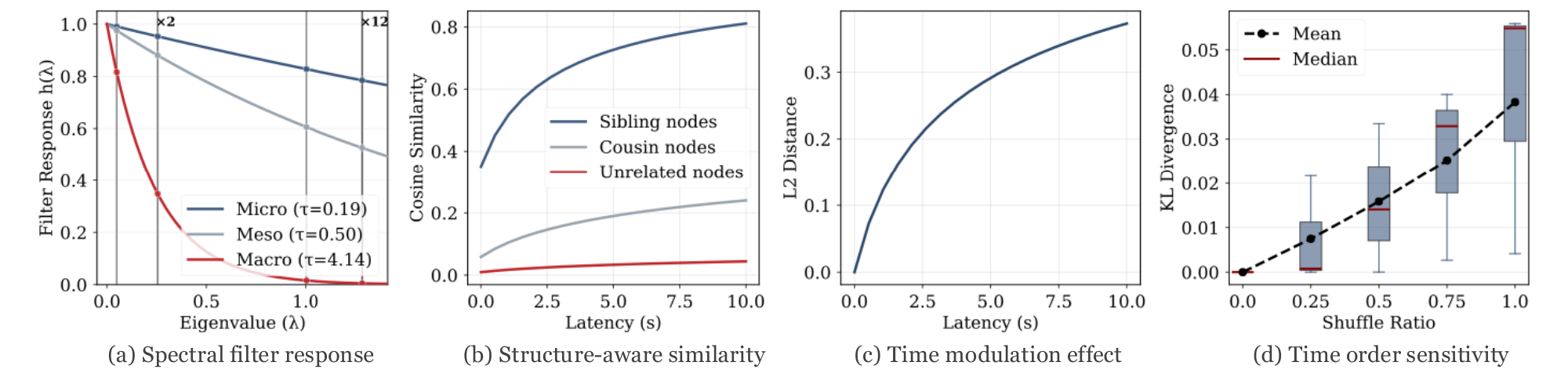}
\end{center}
\vspace{-5pt}
\caption{
\small
(a) Heat kernel filter response at three scales. 
(b) Diffused representations preserve the ultrametric distance in cosine similarity.
(c) Temporal modulation effect: embedding distance increases with latency. 
(d) Time order sensitivity: prediction divergence grows with shuffle ratio.
All panels use a TrajLift model trained on one HKLLT-4 split.
}
\vspace{-10pt}
\label{fig:comb-prec}
\end{figure}

\subsection{Analysis}

We now examine what TrajLift learns with a model trained on HKLLT-4: whether its representations reflect the semantic hierarchy, and how they respond to retrieval timing.

\noindent \textbf{Does TrajLift capture hierarchical structure?}
Figure~\ref{fig:comb-prec} visualizes the learned spectral filters and their effect on node representations.
Panel (a) shows the heat kernel response $h(\lambda) = e^{-\tau\lambda}$ at three scales. The learned diffusion times exhibit clear separation: micro ($\tau$=0.19) preserves high-frequency modes for intra-category distinctions, while macro ($\tau$=4.14) retains only low frequencies capturing cross-domain structure. 
The learned filters thus span the hierarchy, from item-level detail at the micro scale to the domain split at the macro scale.
Panel (b) shows cosine similarity between a node and others at different hierarchical distances. Same-category nodes maintain high similarity, while unrelated nodes remain distant, demonstrating that the learned representations faithfully encode the ultrametric substrate structure.

\noindent \textbf{Is TrajLift sensitive to timing?}
Figure~\ref{fig:comb-prec} examines the model's sensitivity to temporal information.
Panel (c) shows the L2 distance between modulated and unmodulated embeddings as latency varies. The monotonic increase confirms that temporal modulation effectively encodes continuous-valued latencies into the embedding space.
Panel (d) measures prediction sensitivity to temporal order via a shuffle test. We progressively shuffle latency sequences and compute KL divergence from original predictions. The increasing divergence with shuffle ratio demonstrates that the model relies on temporal ordering, not just marginal latency statistics.

\section{Conclusion}
We introduced TrajLift, a spectral framework that models verbal recall as timed trajectories on a semantic hierarchy.
TrajLift lifts each recalled word onto the hierarchy by heat diffusion and lets the latency after each word modulate the diffusion time, so that structure and timing are encoded by a single operator.
Our formal analysis shows that the resulting representation is first-order separable into structural and temporal components, with spectral selectivity across hierarchical levels.
On synthetic benchmarks that require structure-time interaction, TrajLift remains near-perfect on deep hierarchies where all baselines fail.
On real-world clinical data, it outperforms both conventional clinical measures and computational baselines, and its advantage is clearest when retrieval timing is more informative.
More broadly, we believe TrajLift offers a perspective in which clinical knowledge of cognitive impairment could guide representation learning as an inductive bias, as a step toward more interpretable cognitive screening.

\section*{Limitations}
\vspace{-4pt}
The spectral analysis in this work assumes a balanced hierarchical partition, which holds for the HKLLT test protocol used in our experiments. Applications to domains where the semantic hierarchy is unbalanced---for instance, taxonomies with varying branching factors---would require extending the theoretical analysis, although empirically balance appears sufficient rather than necessary (Appendix~\ref{app:synthetic-robustness}). On the empirical side, each ST-XOR variant places the structure-time interaction at a single hierarchical level, so the benchmark does not test whether combining multiple scales is necessary. Clinical datasets for AD screening are inherently small, limiting the statistical power and generalizability of the real-world experiments, and our NCD labels are syndromic rather than specific to AD. Besides, the data used in this study were collected from a local clinical population, and demographic or cultural differences across populations may introduce distributional shifts that we have not evaluated. Validating TrajLift on larger, multi-site cohorts and on other domains involving timed trajectories on hierarchical structures (e.g., navigation in knowledge graphs, structured browsing behavior) is a natural direction for future work.

\section*{Broader Impacts}
\vspace{-4pt}
This work is at the research stage and has not been deployed in any clinical or real-world setting.
This work develops computational tools for early screening of cognitive impairment from verbal memory tests, which are non-invasive, low-cost, and widely administered.
If validated at scale, such tools could support early detection in resource-limited settings where access to neuroimaging or biomarker testing is scarce.
However, the model is intended as a screening aid, not a substitute for clinical diagnosis: false positives may cause unnecessary anxiety, while false negatives may delay intervention.
As noted above, the model has been evaluated on a single local cohort, and deployment across different demographic, cultural, or linguistic populations requires careful re-validation to avoid systematic misdiagnosis.
More broadly, cognitive assessment data is sensitive medical information; safeguards against unauthorized use are essential before any clinical deployment.

\section*{Ethical Statement}
\vspace{-4pt}
The clinical data used in this study were collected under 
a protocol approved by the relevant institutional ethics 
committee.
All participants provided informed consent prior to participation.
All data were de-identified before analysis, and no personally identifiable information is included in this paper or its supplementary materials.

\begin{ack}
\vspace{-4pt}
We thank Dr.~Jingyan Zhou for insightful discussions that shaped the framing and positioning of this work.
This work is supported by the HKSARG Research Grants Council's Theme-based Research Grant Scheme (Project No. T45-407/19N), the CUHK Stanley Ho Big Data Decision Research Centre, and the Centre for Perceptual and Interactive Intelligence (CPII) Ltd, a CUHK-led InnoCentre under the InnoHK initiative of the Innovation and Technology Commission of the Hong Kong Special Administrative Region Government.
\end{ack}

\bibliographystyle{unsrtnat}
\bibliography{main}

\appendix
\newpage
\section{Proof of Proposition~\ref{prop:resolution}}
\label{app:proofs}

We prove Proposition~\ref{prop:resolution} for a balanced two-level hierarchy and then describe how the result extends recursively to deeper balanced hierarchies.

\subsection{Two-Level Case}

\noindent\textbf{Setup.}
Consider $N = Cn$ nodes partitioned into $C$ clusters of size $n$.
We set $W_{ii} = 0$, so weights are assigned only to distinct node pairs: all pairs within the same cluster receive weight $w_1$, and all pairs in different clusters receive weight $w_2$, with $\rho = w_1 / w_2 > 1$.
By the balanced block structure, the graph is regular with degree
\[
    d \;=\; (n-1)w_1 + n(C-1)w_2,
\]
and the normalized Laplacian simplifies to $L = I - W/d$.

\noindent\textbf{Eigenvalue structure.}
The block symmetry of $W$ implies that its eigenspaces decompose into three invariant subspaces.

\emph{Trivial mode} (multiplicity $1$).
The constant vector gives $\lambda_0 = 0$.

\emph{Coarse modes} (multiplicity $C - 1$).
These eigenvectors are constant within each cluster and orthogonal to the constant vector.
For such a vector $\mathbf{u}$, if node $i$ lies in cluster $a$ with cluster-level value $c_a$ and $\sum_{b} c_b = 0$, then
\[
    (W\mathbf{u})_i
    \;=\;
    (n-1)w_1 \, c_a + n w_2 \sum_{b \neq a} c_b
    \;=\;
    \big((n-1)w_1 - nw_2\big) \, c_a.
\]
The corresponding Laplacian eigenvalue is
\[
    \lambda_{\mathrm{coarse}}
    \;=\;
    1 - \frac{(n-1)w_1 - nw_2}{d}
    \;=\;
    \frac{nCw_2}{d}.
\]

\emph{Fine modes} (multiplicity $C(n-1)$).
These eigenvectors sum to zero within each cluster.
For such a vector, inter-cluster contributions vanish (since each cluster's entries sum to zero), and within-cluster contributions satisfy $(W\mathbf{u})_i = -w_1 u_i$.
The corresponding Laplacian eigenvalue is
\[
    \lambda_{\mathrm{fine}}
    \;=\;
    1 + \frac{w_1}{d}
    \;=\;
    \frac{n w_1 + n(C-1)w_2}{d}.
\]

\noindent\textbf{Spectral gap.}
Writing the eigenvalues in terms of $\rho = w_1/w_2$:
\[
    \lambda_{\mathrm{coarse}}
    \;=\;
    \frac{nC}{(n-1)\rho + n(C-1)},
    \qquad
    \lambda_{\mathrm{fine}}
    \;=\;
    \frac{n(\rho + C - 1)}{(n-1)\rho + n(C-1)}.
\]
The gap between the two non-trivial eigenspaces is
\[
    \gamma
    \;=\;
    \lambda_{\mathrm{fine}} - \lambda_{\mathrm{coarse}}
    \;=\;
    \frac{n(\rho - 1)}{(n-1)\rho + n(C-1)}.
\]
When $\rho \gg C$, the coarse eigenvalue is $O(C/\rho)$, while the fine eigenvalue and the gap are both $O(1)$.
This confirms that for sufficiently large weight ratios, the Laplacian spectrum separates into two distinct bands corresponding to coarse (cluster-level) and fine (within-cluster) structure.

\noindent\textbf{Heat kernel behavior.}
The heat kernel $H_\tau = e^{-\tau L}$ attenuates each eigenmode by $e^{-\tau \lambda_i}$.
When $\rho \gg C$, there exists a non-empty intermediate range $1 \ll \tau \ll \rho/C$, and the two non-trivial bands respond differently:
\begin{itemize}
    \item Coarse modes: attenuated by $e^{-\tau \lambda_{\mathrm{coarse}}}$.
    Since $\lambda_{\mathrm{coarse}} = O(C/\rho)$, this factor remains close to $1$ as long as $\tau \ll \rho/C$.

    \item Fine modes: attenuated by $e^{-\tau \lambda_{\mathrm{fine}}}$.
    Since $\lambda_{\mathrm{fine}} = O(1)$, this factor becomes small when $\tau \gg 1$.
\end{itemize}
This yields three regimes of the diffusion time:
\begin{enumerate}
    \item $\tau \ll 1$: both bands are preserved, and the lifted representation retains full within-cluster detail.
    \item $1 \ll \tau \ll \rho/C$: fine modes are suppressed while coarse modes persist, yielding a cluster-level representation.
    \item $\tau \gg \rho/C$: all non-trivial modes are suppressed, and the representation collapses to the uniform graph mean (which is the unique stationary component on a regular graph).
\end{enumerate}
Therefore, increasing $\tau$ continuously interpolates from item-level to cluster-level representations, with the transition governed by the spectral gap $\gamma$. \qed

\subsection{Multi-Level Extension}

For a balanced nested hierarchy in which all groups at the same level have equal size and edge weights depend only on the ultrametric distance (Definition~\ref{def:ultrametric}), the partition at each level is equitable.
The subspaces of vectors that are constant on blocks at one level and sum to zero within their parent block are invariant under $W$ and hence under $L$.
The exact eigenvalues in the multi-level case depend on the branching factors and all weights across levels.
Nevertheless, applying the two-level argument to each adjacent refinement shows that, when adjacent-level ratios $\rho_\ell = w(\ell) / w(\ell+1)$ are sufficiently separated, these invariant subspaces form spectral bands associated with successive hierarchy refinements, together with the trivial zero eigenvalue.
Heat diffusion then suppresses these bands from fine to coarse as $\tau$ increases.

In our experimental setting, the HKLLT hierarchy is balanced: 2 domains $\times$ 2 categories per domain $\times$ 4 items per category, with weight ratios $w(1)/w(2) = 10$ and $w(2)/w(3) = 10$.
These ratios are chosen to encourage separated spectral bands, which we target with the micro, meso, and macro diffusion scales defined in \S\ref{sec:lifting}.

\section{Robustness on Synthetic Benchmarks}
\label{app:synthetic-robustness}

We examine three aspects of the ST-XOR-wh results: hierarchy balance, timestamp noise, and the choice of diffusion scales.

\subsection{Unbalanced Hierarchies}
Proposition~1 and Appendix~A assume a balanced partition. We relax this assumption on ST-XOR-wh by drawing cluster sizes unevenly at a fixed total number of nodes. As Table~\ref{tab:unbalanced} shows, TrajLift's accuracy remains around 99\% as imbalance grows. Balance thus appears sufficient rather than necessary.

\begin{table}[h]
\centering
\caption{Unbalanced hierarchies on ST-XOR-wh. CV: coefficient of variation of cluster sizes.}
\label{tab:unbalanced}
\begin{tabular}{lcccc}
\toprule
CV of cluster size & 0 & 0.232 & 0.524 & 0.970 \\
\midrule
TrajLift accuracy (\%) & 98.67$\pm$0.57 & 98.67$\pm$1.18 & 98.87$\pm$0.61 & 99.00$\pm$0.41 \\
\bottomrule
\end{tabular}
\end{table}

\subsection{Corrupted Timestamps}
We perturb latencies multiplicatively at test time, $\delta \to \delta\varepsilon$ with $\log \varepsilon \sim \mathcal{N}(0, \sigma^2)$. To express noise strength on an interpretable scale, we report the discriminability of the fast and slow latency distributions that define the two classes, $d'(\sigma) = 1.80 / \sqrt{0.184^2 + \sigma^2}$, where $1.80$ is their separation and $0.184$ their pooled standard deviation on the log scale, together with their overlap $2\Phi(-d'/2)$. As Table~\ref{tab:timestamp} shows, TrajLift retains 95\% accuracy when the two distributions overlap by about a quarter, and approaches chance only when $d' < 1$, where the distributions themselves are barely separable.

\begin{table}[h]
\centering
\caption{Test-time timestamp perturbation on ST-XOR-wh.}
\label{tab:timestamp}
\begin{tabular}{lccccccc}
\toprule
$\sigma$ & 0 & 0.25 & 0.50 & 0.75 & 1.00 & 1.50 & 2.00 \\
\midrule
$d'$ & 9.77 & 5.79 & 3.37 & 2.33 & 1.77 & 1.19 & 0.90 \\
Overlap (\%) & 0 & 0.4 & 9.2 & 24.4 & 37.7 & 55.2 & 65.4 \\
TrajLift accuracy (\%) & 98.67 & 99.13 & 97.87 & 95.00 & 79.40 & 60.80 & 53.53 \\
\bottomrule
\end{tabular}
\end{table}

\subsection{Choice of Diffusion Scales}
ST-XOR-wh has three hierarchy levels, matching the three diffusion scales of TrajLift. To test whether this match drives the result, we train TrajLift with each of the seven non-empty subsets of \{micro, meso, macro\}. As Table~\ref{tab:scales} shows, every subset containing the macro scale reaches about 99\%, regardless of the number of scales, whereas subsets without it fail or degrade. Performance thus depends on whether a scale resolves the level that carries the signal, not on the number of scales. 
With \{micro\} alone, the heat kernel is close to the identity ($\tau \approx 0.12$), and accuracy falls to the level of the structure-agnostic LSTM$^{+}$ (Table~\ref{tab:syn}).

\begin{table}[h]
\centering
\caption{TrajLift with subsets of diffusion scales on ST-XOR-wh (accuracy, \%), mean $\pm$ SD over five random seeds.}
\label{tab:scales}
\begin{tabular}{lcc}
\toprule
Scales & Number & Accuracy \\
\midrule
\{macro\} & 1 & 99.67$\pm$0.75 \\
\{meso, macro\} & 2 & 99.47$\pm$0.30 \\
\{micro, macro\} & 2 & 99.27$\pm$0.60 \\
\{micro, meso, macro\} & 3 & 98.67$\pm$0.57 \\
\{meso\} & 1 & 94.47$\pm$3.74 \\
\{micro, meso\} & 2 & 53.20$\pm$2.33 \\
\{micro\} & 1 & 52.73$\pm$3.93 \\
\bottomrule
\end{tabular}
\end{table}

\section{Data Details}
\label{app:data}

\subsection{Hong Kong List Learning Test (HKLLT)}
\label{app:hkllt}

The Hong Kong List Learning Test (HKLLT)~\citep{chan1998hkllt} is a standardized neuropsychological assessment designed for Chinese-speaking populations to evaluate verbal learning and episodic memory. Episodic memory impairment is among the earliest cognitive deficits observed in Alzheimer's disease, reflecting neuropathological changes in the hippocampus and entorhinal cortex that occur in the initial stages of the disease~\citep{braak1991neuropathological}. As the disease progresses, the frontal, temporal, and parietal association cortices become increasingly involved, leading to broader deficits in lexical-semantic functions~\citep{au2003verbal}. The HKLLT is specifically designed to probe these capacities through the recall of a 16-word list organized under a three-level semantic hierarchy, as detailed in Table~\ref{tab:wordlist}.

\begin{table}[htbp]
\centering
\caption{HKLLT word list and semantic hierarchy. The 16 target words are organized into four categories under two domains.}
\label{tab:wordlist}
\begin{tabular}{lll}
\toprule
\textbf{Domain} & \textbf{Category} & \textbf{Items} \\
\midrule
\multirow{2}{*}{Abstract} 
    & Relatives & Grandmother, Aunt, Niece, Cousin \\
    & Countries & India, Switzerland, Laos, Chile \\
\midrule
\multirow{2}{*}{Concrete} 
    & Vegetables & Eggplant, Cucumber, Mustard Greens, Onion \\
    & Furniture & Desk, Mirror, Lamp, Wardrobe \\
\bottomrule
\end{tabular}
\end{table}

The full protocol comprises five trials: Trial~1 presents the 16 words orally in randomized order for immediate free recall; Trials~2--3 repeat the same list to measure learning rate; Trial~4 assesses delayed recall after a 10-minute non-verbal interval; Trial~5 assesses delayed recall after 30 minutes; a final recognition phase tests discrimination between targets and distractors.

In this work, we focus exclusively on Trial~4 and Trial~5. Clinical research has established that delayed recall is particularly effective for early-stage AD detection, as the rate of forgetting over these intervals is the most sensitive discriminator between patients with mild cognitive impairment and cognitively normal elderly~\citep{au2003verbal}.

\subsection{Participant Demographics}
\label{app:demographics}

Table~\ref{tab:demographics} summarizes participant characteristics. The NC and NCD groups differ in age and education: the NCD group is older and has fewer years of education (Welch's $t$-test, NC vs.\ NCD: $t = -4.84$ for age and $t = 3.67$ for education; both $p < 0.001$), consistent with known epidemiological patterns. TrajLift receives no demographic input, and the low-level feature baseline (\S5.1) includes both variables.

\begin{table}[h]
\centering
\caption{Demographic characteristics of participants. Continuous variables are reported as mean $\pm$ SD; gender as Male/Female counts.}
\label{tab:demographics}
\begin{tabular}{lcc}
\toprule
 & NC & NCD \\
\midrule
Number & 233 & 154 \\
Age (years) & 70.8 $\pm$ 6.5 & 74.5 $\pm$ 7.8 \\
Education (years) & 8.8 $\pm$ 4.1 & 7.2 $\pm$ 4.4 \\
Gender (M/F) & 98/135 & 74/80 \\
\bottomrule
\end{tabular}
\end{table}

\subsection{Human Subjects and Data Collection Protocol}
The clinical cohort was collected under a protocol approved by the institutional research ethics committee. Data were collected by trained personnel following the standard HKLLT administration procedure. All participants provided informed consent and were compensated for their participation. The data contain protected health information and are not publicly released.

\subsection{Measure Validation}
\label{app:measure_validation}

To validate that our dataset exhibits expected patterns, we examine three commonly used measures via independent-samples $t$-tests (Table~\ref{tab:measure_validation}):

\begin{itemize}[leftmargin=*,nosep]
\item \textbf{Semantic Clustering:} number of consecutive same-category recalls, reflecting sensitivity to hierarchical organization of semantic memory.
\item \textbf{Recall Count:} total correctly recalled words (max 16), indexing overall retention capacity.
\item \textbf{Recency Effect:} proportion of last-four-list items among recalls, measuring serial position effects.
\end{itemize}

\begin{table}[htbp]
\centering
\caption{Measure validation via group comparison ($t$-test). Values are mean $\pm$ SD.}
\label{tab:measure_validation}
\begin{tabular}{llcccc}
\toprule
\textbf{Measure} & \textbf{Trial} & \textbf{NC} & \textbf{NCD} & $t$ & $p$ \\
\midrule
\multirow{2}{*}{Semantic Clustering} 
    & T4 & $3.33 \pm 2.99$ & $1.38 \pm 2.23$ & 6.91 & $<.001$ \\
    & T5 & $2.99 \pm 2.61$ & $1.02 \pm 1.47$ & 8.48 & $<.001$ \\
\midrule
\multirow{2}{*}{Recall Count} 
    & T4 & $6.82 \pm 2.50$ & $3.21 \pm 2.50$ & 13.87 & $<.001$ \\
    & T5 & $6.36 \pm 2.62$ & $2.69 \pm 2.25$ & 14.18 & $<.001$ \\
\midrule
\multirow{2}{*}{Recency Effect} 
    & T4 & $0.12 \pm 0.12$ & $0.12 \pm 0.20$ & $-0.02$ & .984 \\
    & T5 & $0.12 \pm 0.11$ & $0.13 \pm 0.22$ & $-0.65$ & .517 \\
\bottomrule
\end{tabular}
\end{table}

Both semantic clustering and recall count show significant group differences ($p < 0.001$), consistent with clinical findings that patients with AD recall fewer words and exhibit reduced categorical organization. 
Recency effects show no significant difference ($p > 0.5$), as expected in delayed recall, where the recency advantage dissipates.

\subsection{Trajectory Statistics}
\label{app:trajectory_stats}

Table~12 reports trajectory-level statistics by trial and group, computed per participant and averaged within each group. Specifically, sequence length counts all recalled items, including repetitions; mean inter-item latency averages the latencies $\delta_k$ within a trajectory; category transitions counts consecutive recalls from different categories, complementing semantic clustering (Appendix~C.4); category coverage counts the distinct categories recalled; word coverage counts the distinct words recalled.

\begin{table}[htbp]
\centering
\caption{Trajectory statistics by trial and group. Values are mean $\pm$ SD.}
\label{tab:traj_stats}
\begin{tabular}{llcc}
\toprule
\textbf{Statistic} & \textbf{Trial} & \textbf{NC} & \textbf{NCD} \\
\midrule
\multirow{2}{*}{Sequence Length} 
    & T4 & $9.06 \pm 4.60$ & $4.52 \pm 4.24$ \\
    & T5 & $7.76 \pm 3.86$ & $3.42 \pm 3.06$ \\
\midrule
\multirow{2}{*}{Mean Inter-Item Latency (s)} 
    & T4 & $2.70 \pm 1.38$ & $3.27 \pm 2.14$\\
    & T5 & $2.39 \pm 1.25$ & $3.49 \pm 2.08$ \\
\midrule
\multirow{2}{*}{Word Coverage} 
    & T4 & $6.82 \pm 2.50$ & $3.21 \pm 2.50$ \\
    & T5 & $6.36 \pm 2.62$ & $2.69 \pm 2.25$ \\
\midrule
\multirow{2}{*}{Category Coverage} 
    & T4 & $3.57 \pm 0.65$ & $2.21 \pm 1.40$ \\
    & T5 & $3.40 \pm 0.83$ & $1.94 \pm 1.38$ \\
\midrule
\multirow{2}{*}{Category Transitions} 
    & T4 & $4.73 \pm 2.52$ & $2.34 \pm 2.37$ \\
    & T5 & $3.79 \pm 1.99$ & $1.65 \pm 1.84$ \\
\bottomrule
\end{tabular}
\end{table}
\section{Implementation Details}
\label{appendix:implementation}

\subsection{TrajLift Configurations}
\subsubsection{Diffusion Time Initialization}
 
The base diffusion time at each scale is initialized from the largest eigenvalue of $L$:
\begin{equation}
    \tau_s^{\mathrm{base}} \;=\; \frac{-\ln(\alpha_s)}{\lambda_{\max}},
    \label{eq:tau_init}
\end{equation}
where $\alpha_s \in (0,1)$ is a target retention rate controlling how much of the highest-frequency component survives at scale $s$.
We set $\alpha_{\mathrm{micro}} = 0.85$, $\alpha_{\mathrm{meso}} = 0.25$, and $\alpha_{\mathrm{macro}} = 0.002$, corresponding to local, community-level, and global diffusion respectively.
High retention (micro) preserves within-cluster fine structure, while low retention (macro) suppresses all but the coarsest modes.
 
\subsubsection{Modulation Coefficient Initialization}
 
The modulation coefficient $\gamma_s$ in Equation~\eqref{eq:modulation} controls the range of the latency-induced perturbation $\Delta\tau_k^{(s)} = \tau_s^{\mathrm{base}} \cdot \gamma_s \cdot \varphi(\delta_k)$.
We initialize $\gamma_s$ from training set latency statistics:
\begin{equation}
    \gamma_s \;=\; \frac{r - 1}{\varphi(\delta_{90})},
    \label{eq:gamma_init}
\end{equation}
where $r = 3.0$ is a target variation ratio and $\delta_{90}$ is the 90th percentile of inter-item latencies in the training set.
This ensures that for the large majority of observed latencies, the modulated diffusion time $\tilde{\tau}_k^{(s)}$ stays within a factor of $r$ of the base time: $\tilde{\tau}_k^{(s)} \le r \cdot \tau_s^{\mathrm{base}}$ for $\delta_k \le \delta_{90}$.
We apply $\varphi(\delta) = \log(1+\delta)$ with latencies clipped at 20 seconds to prevent extreme values from producing excessive modulation.

\begin{figure}[h]
\begin{center}
\includegraphics[width=0.85\linewidth,scale=1.0]{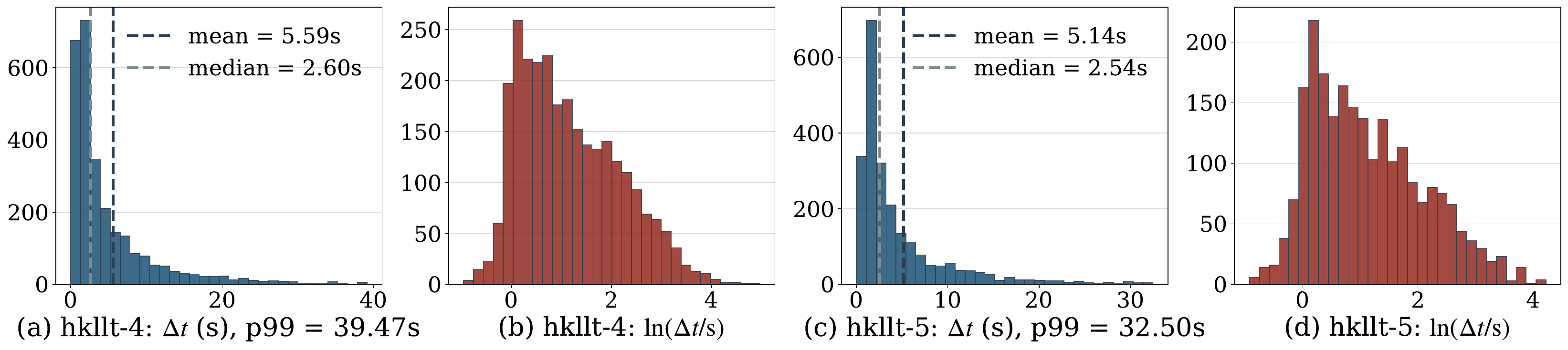}
\end{center}
\caption{
\small
Distribution of inter-item latencies in the HKLLT dataset. (a,c) Raw latencies $\delta_k$ for Trials 4 and 5: the distributions are heavily right-skewed (mean $\gg$ median) with 99th percentiles exceeding 30s. (b,d) Log-transformed latencies $\ln(\delta_k)$: the distributions become approximately symmetric. Latencies exceeding 20s are clipped before transformation.
}
\label{fig:latency_dist}
\end{figure}

\noindent\textbf{Choice of $\varphi$.}
Figure~\ref{fig:latency_dist} shows the distribution of inter-item latencies in the HKLLT data.
The raw latencies are heavily right-skewed (mean $\approx$ 5.6s versus median $\approx$ 2.6s for Trial~4, with the 99th percentile exceeding 39s), so a linear mapping would allow rare extreme pauses to dominate the modulation.
After a log transform, the distribution becomes approximately symmetric, so $\varphi$ concentrates representational sensitivity on the frequent short-to-moderate latency range while compressing the tail.

\subsubsection{Spectral Gap Margin Verification}
\label{app:margin}

We verify the range of the latency-induced modulation claimed in \S\ref{sec:temporal}.

From the eigenvalue analysis in Appendix~\ref{app:proofs}, the spectral gap between the fine and coarse eigenvalue bands of our HKLLT graph is
\[
    \gamma \;=\; \frac{n(\rho - 1)}{(n-1)\rho + n(C-1)},
\]
where $n$ is the cluster size, $C$ is the number of clusters, and $\rho$ is the weight ratio at the relevant hierarchy level.
For the category level ($n = 4$, $C = 4$, $\rho = 10$), this gives $\gamma = \frac{4 \times 9}{3 \times 10 + 4 \times 3} = \frac{36}{42} \approx 0.86$.
The inverse spectral gap $1/\gamma \approx 1.17$ sets the scale at which modulation would begin to cross eigenvalue bands.
 
The maximum modulation at scale $s$ is
\[
    \Delta\tau_{\max}^{(s)} \;=\; \tau_s^{\mathrm{base}} \cdot \gamma_s \cdot \varphi(\delta_{90}).
\]
By construction (Equation~\eqref{eq:gamma_init}), $\gamma_s \cdot \varphi(\delta_{90}) = r - 1 = 2.0$, so $\Delta\tau_{\max}^{(s)} = 2 \, \tau_s^{\mathrm{base}}$.
 
For the micro scale, $\tau_{\mathrm{micro}}^{\mathrm{base}} = -\ln(0.85)/\lambda_{\max}$.
With $\lambda_{\max} \approx n/(n-1) = 4/3$ (from Appendix~\ref{app:proofs}), we get $\tau_{\mathrm{micro}}^{\mathrm{base}} \approx 0.12$, and therefore $\Delta\tau_{\max}^{(\mathrm{micro})} \approx 0.24$.
Since $0.24 \ll 1/\gamma \approx 1.17$, the micro-scale modulation stays well within the spectral gap margin.
For the meso and macro scales, $\Delta\tau_{\max}$ exceeds this margin; since the modulated diffusion time is bounded by $3\,\tau_s^{\mathrm{base}}$ and adjacent base times differ by more than a factor of three, modulation still does not move one scale into the range of another.

\subsubsection{Band Separation of Multi-Scale Residuals}
 
Proposition~\ref{prop:bandpass} establishes that the temporal residual at scale $s$ has a band-pass-shaped spectral response peaking at $\lambda^\star = 1/\tau_s^{\mathrm{base}}$.
For the three scales to capture non-redundant temporal information, these peaks should fall in different eigenvalue bands.
 
From the initialization above, the peak frequencies are:
\[
    \frac{1}{\tau_{\mathrm{micro}}^{\mathrm{base}}} \;=\; \frac{\lambda_{\max}}{-\ln(0.85)} \;\approx\; 8.2,
    \qquad
    \frac{1}{\tau_{\mathrm{meso}}^{\mathrm{base}}} \;\approx\; 0.96,
    \qquad
    \frac{1}{\tau_{\mathrm{macro}}^{\mathrm{base}}} \;\approx\; 0.21.
\]
In our HKLLT graph, the fine eigenvalues are near $\lambda_{\mathrm{fine}} \approx n/(n-1) \approx 1.33$ and the coarse eigenvalues are near $\lambda_{\mathrm{coarse}} \approx C/\rho \approx 0.4$.
The micro peak ($\approx 8.2$) lies above the fine band, the meso peak ($\approx 0.96$) lies near the fine band, and the macro peak ($\approx 0.21$) lies below the coarse band.
These peaks therefore target different spectral regions, confirming that the three temporal residual channels capture temporal information at distinct hierarchical resolutions.

\subsubsection{Architecture}
The sequence encoder is a 2-layer LSTM with a hidden dimension of 16 and a dropout rate of 0.1. While other sequential architectures, such as Transformers, are compatible with our framework, we adopt LSTM due to its robustness under limited clinical data.
After LSTM, we use a linear classifier for classification.

\subsection{Baseline Configurations}

\subsubsection{Shared Settings}
All baselines embed each node into 8 dimensions via one-hot encoding followed by linear projection.
The hidden dimension is set to 16 across all models. For time-augmented variants (denoted with `+'), we use a Time2Vec dimension of 4. All other training configurations (learning rate, optimizer, etc.) are identical to the proposed method.

\subsubsection{Trajectory Learning Baselines}
Table~\ref{tab:baseline_config} summarizes the configuration for each baseline model. All models use a dropout rate of 0.1.

\begin{table}[h]
\centering
\caption{Baseline model configurations.}
\label{tab:baseline_config}
\begin{tabular}{ll}
\toprule
Model & Configuration \\
\midrule
LSTM$^{+}$ & 2 layers \\
Transformer$^{+}$ & 2 layers, 2 attention heads, feed-forward dimension 32 \\
TGN & memory dimension 16, message dimension 16 \\
TGAT & 2 attention layers, 2 attention heads \\
TrajGAT$^{+}$ & 2 GAT layers, 2 attention heads \\
Graphormer$^{+}$ & 2 layers, 2 attention heads, feed-forward dimension 32 \\
GAT$^{+}$ w/ LSTM & 2 GAT layers, 2 attention heads, 1-layer LSTM \\
R-GCN$^{+}$ w/ LSTM & 2 GNN layers, 2 relation bases, 1-layer LSTM \\
\bottomrule
\end{tabular}
\end{table}

\subsubsection{NCD Detection Baselines}
\label{app:ncd_detection_baselines}
For PLM baselines, we use Chinese BERT\footnote{\url{https://huggingface.co/google-bert/bert-base-chinese}} and Chinese RoBERTa\footnote{\url{https://huggingface.co/hfl/chinese-roberta-wwm-ext}} with frozen encoders. We train a classifier with a hidden dimension of 16 and a dropout rate of 0.1 on [CLS] token representations. 
For handcrafted baselines, we train logistic regression on the measures listed in \S5.1 (defined in Appendix~C). Adding all 21 pairwise interactions to the low-level feature set (28-D) yields 74.53 and 75.56 accuracy on HKLLT-4 and HKLLT-5, below the 7-D model.

\subsubsection{Graph Inputs}
This subsection clarifies the graph inputs used by different trajectory-learning baselines. The hierarchy-edge construction applies only to the static-graph baselines in categories (c) and (d) of \S5.1, i.e., TrajGAT+, Graphormer+, GAT+ w/ LSTM, and R-GCN+ w/ LSTM. It does not apply to the dynamic graph baselines TGN and TGAT, which are evaluated in their native continuous-time dynamic-graph setting using temporal item-to-item events.

For each recall trajectory used by the static-graph baselines, we construct a trajectory-specific graph. Item nodes correspond to the unique recalled items in the trajectory. Temporal transition edges connect consecutive recalled items in recall order and carry the inter-item latency encoded by Time2Vec. To expose the same semantic hierarchy used by TrajLift, we additionally include category and domain nodes and add hierarchy edges from each item node to its parent category node and parent domain node. Thus, these baseline graphs contain two edge types: temporal transition edges and semantic hierarchy edges.

The representation of edge types follows each baseline’s native interface. For R-GCN+ w/ LSTM, temporal transition edges and hierarchy edges are encoded as separate relation types. For GAT+ w/ LSTM and TrajGAT+, both edge types are included in the same message-passing graph; edge attributes distinguish temporal transition edges from hierarchy edges, with Time2Vec latency features used for transition edges and a zero-latency/type indicator used for hierarchy edges. For Graphormer+, the combined graph is used to compute structural encodings, and edge-type information is included through edge embeddings distinguishing transition and hierarchy edges. No hierarchy edges are added to TGN or TGAT because these models represent the input as a stream of timestamped interaction events rather than as message passing over a fixed semantic hierarchy.

\subsection{Training Protocol}

All models are trained using the Adam optimizer with a learning rate of 0.001 and weight decay of 0.0001. The batch size is set to 16. For synthetic data, we apply early stopping with a patience of 20 epochs based on validation accuracy, up to a maximum of 100 epochs. For the HKLLT data, we set the training epochs to 50. To address class imbalance, class weights are computed using inverse frequency. All experiments are implemented in PyTorch 2.1.1 and PyTorch Geometric 2.6.1.

\subsection{Existing Assets and Licenses}
We use the following existing assets. For pretrained language-model baselines, we use bert-base-chinese from Google and chinese-roberta-wwm-ext from HFL via HuggingFace. These models are used only as frozen encoders for baseline comparisons. 
We cite the corresponding model sources and respect the licenses and terms of use distributed with the HuggingFace repositories (Appendix~\ref{app:ncd_detection_baselines}). 
For implementation, we use PyTorch 2.1.1+cu118 and PyTorch Geometric 2.6.1, which are released under BSD-style licenses. For graph and sequence baselines, we implement the compared architectures following their original papers and cite the corresponding publications; no pretrained checkpoints from these baselines are used. The HKLLT clinical dataset is proprietary clinical data used under an IRB-approved protocol and confidentiality agreements; it is not redistributed or repackaged. The synthetic ST-XOR benchmarks are generated by the authors and are released with the code under the license specified in the repository.

\subsection{Compute Resources}
All experiments were conducted on RTX8000 with 48 GB GPU memory using PyTorch 2.1.1+cu118 and PyTorch Geometric 2.6.1. 
A single training run of TrajLift takes approximately 2-8 minutes on each synthetic benchmark and 2-6 minutes on each HKLLT trial.
\newpage
\section{Sample Visualization}
\label{appendix:sample}

\begin{figure*}[h!]
\begin{center}
\includegraphics[width=0.8\linewidth,scale=1.0]{Figures/sample-4-final.pdf}
\end{center}
\vspace{-10pt}
\caption{
Example recall trajectories from HKLLT Trial 4 (Class 1 = NCD, Class 0 = NC). Words are laid out by category. Arrows follow recall order and are labeled with the inter-item latency in seconds.
}
\vspace{-20pt}
\label{fig:samples4}
\end{figure*}

\newpage
\begin{figure*}[h!]
\begin{center}
\includegraphics[width=0.8\linewidth,scale=1.0]{Figures/sample-5-final.pdf}
\end{center}
\vspace{-10pt}
\caption{
Example recall trajectories from HKLLT Trial 5 (Class 1 = NCD, Class 0 = NC). Words are laid out by category. Arrows follow recall order and are labeled with the inter-item latency in seconds.
}
\vspace{-20pt}
\label{fig:samples5}
\end{figure*}

\newpage

\newpage
\section*{NeurIPS Paper Checklist}

\begin{enumerate}

\item {\bf Claims}
    \item[] Question: Do the main claims made in the abstract and introduction accurately reflect the paper's contributions and scope?
    \item[] Answer: \answerYes{} 
    \item[] Justification: The claims in the abstract and introduction are aligned with the proposed TrajLift framework, the theoretical analysis, and the empirical evaluations on synthetic benchmarks and a clinical HKLLT dataset. We also added the theoretical limitation statement in the Limitation section.

\item {\bf Limitations}
    \item[] Question: Does the paper discuss the limitations of the work performed by the authors?
    \item[] Answer: \answerYes{} 
    \item[] Justification: A Limitation section was included to discuss the limitations of this work.

\item {\bf Theory assumptions and proofs}
    \item[] Question: For each theoretical result, does the paper provide the full set of assumptions and a complete (and correct) proof?
    \item[] Answer: \answerYes{} 
    \item[] Justification: The paper states theoretical results as propositions/lemmas and provides formal analysis in the appendix.

    \item {\bf Experimental result reproducibility}
    \item[] Question: Does the paper fully disclose all the information needed to reproduce the main experimental results of the paper to the extent that it affects the main claims and/or conclusions of the paper (regardless of whether the code and data are provided or not)?
    \item[] Answer: \answerYes{} 
    \item[] Justification: We provide model architecture, data construction, preprocessing, train/validation/test splits, hyperparameter selection, baseline configurations, and evaluation protocols. The clinical data contain protected health information and cannot be released, but we provide detailed statistics and release the synthetic benchmark and code at https://github.com/kjw11/TrajLift.

\item {\bf Open access to data and code}
    \item[] Question: Does the paper provide open access to the data and code, with sufficient instructions to faithfully reproduce the main experimental results, as described in supplemental material?
    \item[] Answer: \answerNo{} 
    \item[] Justification: The clinical data contain protected health information and cannot be released. The synthetic benchmark, data-generation code, model implementation, and experimental configurations are released at https://github.com/kjw11/TrajLift.

\item {\bf Experimental setting/details}
    \item[] Question: Does the paper specify all the training and test details (e.g., data splits, hyperparameters, how they were chosen, type of optimizer) necessary to understand the results?
    \item[] Answer: \answerYes{} 
    \item[] Justification: The paper specifies data splits, metrics, validation protocol, hyperparameter selection, model configurations, optimizer, learning rate, batch size, number of epochs, etc., in the experimental section and appendix.

\item {\bf Experiment statistical significance}
    \item[] Question: Does the paper report error bars suitably and correctly defined or other appropriate information about the statistical significance of the experiments?
    \item[] Answer: \answerYes{} 
    \item[] Justification: Synthetic experiments are repeated over five random seeds and HKLLT experiments over 15 stratified splits; tables report mean and standard deviation. Paired comparisons on HKLLT report Nadeau–Bengio corrected 95\% confidence intervals.

\item {\bf Experiments compute resources}
    \item[] Question: For each experiment, does the paper provide sufficient information on the computer resources (type of compute workers, memory, time of execution) needed to reproduce the experiments?
    \item[] Answer: \answerYes{} 
    \item[] Justification: We report the compute environment, including GPU/CPU type, memory, and approximate runtime per experiment for the reported experiments. The total compute used is hard to estimate due to server load fluctuation.
    
\item {\bf Code of ethics}
    \item[] Question: Does the research conducted in the paper conform, in every respect, with the NeurIPS Code of Ethics \url{https://neurips.cc/public/EthicsGuidelines}?
    \item[] Answer: \answerYes{} 
    \item[] Justification: The study follows the NeurIPS Code of Ethics. The clinical data are de-identified and handled under privacy and confidentiality requirements.

\item {\bf Broader impacts}
    \item[] Question: Does the paper discuss both potential positive societal impacts and negative societal impacts of the work performed?
    \item[] Answer: \answerYes{} 
    \item[] Justification: We discuss potential positive impacts for scalable auxiliary cognitive screening as well as risks, including incorrect predictions, and privacy concerns in the Broader Impacts section.
    
\item {\bf Safeguards}
    \item[] Question: Does the paper describe safeguards that have been put in place for responsible release of data or models that have a high risk for misuse (e.g., pre-trained language models, image generators, or scraped datasets)?
    \item[] Answer: \answerYes{} 
    \item[] Justification: We do not release protected clinical data. We state that TrajLift is intended only as an auxiliary screening research tool and should not be used for standalone clinical diagnosis without clinician oversight and external validation.

\item {\bf Licenses for existing assets}
    \item[] Question: Are the creators or original owners of assets (e.g., code, data, models), used in the paper, properly credited and are the license and terms of use explicitly mentioned and properly respected?
    \item[] Answer: \answerYes{} 
    \item[] Justification: We cite and document the existing assets used in the paper, including pretrained BERT/RoBERTa models, software libraries, and the HKLLT protocol, and provide license or terms-of-use information where applicable.

\item {\bf New assets}
    \item[] Question: Are new assets introduced in the paper well documented and is the documentation provided alongside the assets?
    \item[] Answer: \answerYes{} 
    \item[] Justification: The synthetic benchmark, generation scripts, and model code are released at https://github.com/kjw11/TrajLift with documentation in the repository README.

\item {\bf Crowdsourcing and research with human subjects}
    \item[] Question: For crowdsourcing experiments and research with human subjects, does the paper include the full text of instructions given to participants and screenshots, if applicable, as well as details about compensation (if any)? 
    \item[] Answer: \answerYes{} 
    \item[] Justification: The study does not involve crowdsourcing. For the clinical data, we describe the HKLLT protocol, inclusion criteria, participant demographics, and data collection procedure.

\item {\bf Institutional review board (IRB) approvals or equivalent for research with human subjects}
    \item[] Question: Does the paper describe potential risks incurred by study participants, whether such risks were disclosed to the subjects, and whether Institutional Review Board (IRB) approvals (or an equivalent approval/review based on the requirements of your country or institution) were obtained?
    \item[] Answer: \answerYes{}  
    \item[] Justification: The clinical data collection and use of the HKLLT dataset were reviewed and approved by the institutional research ethics committee. The paper describes privacy protection, protected health information constraints, and clinical-use risks.

\item {\bf Declaration of LLM usage}
    \item[] Question: Does the paper describe the usage of LLMs if it is an important, original, or non-standard component of the core methods in this research? Note that if the LLM is used only for writing, editing, or formatting purposes and does \emph{not} impact the core methodology, scientific rigor, or originality of the research, declaration is not required.
    \item[] Answer: \answerNA{} 
    \item[] Justification: LLMs were not used as an important, original, or non-standard component of the core method. They were used only for language polishing, and no proprietary or sensitive clinical data were input into LLMs.

\end{enumerate}

\end{document}